\documentclass[notitlepage]{report}

\usepackage[ruled,norelsize]{algorithm2e}
\usepackage{mathtools}
\usepackage{graphicx}
\usepackage{epstopdf}
\usepackage{amsfonts}
\usepackage{amsmath, amsfonts, amssymb, bm}
\usepackage{url}
\usepackage[latin1]{inputenc}
\usepackage{tikz}
\usetikzlibrary{shapes,arrows}

\title{Hierarchical Manifold Clustering on Diffusion Maps for Connectomics \\ \  \\ \large 18.S096 final project}
\author{Gergely Odor}

\date{}

\begin{document}

\maketitle

\begin{abstract}
In this paper, we introduce a novel algorithm for segmentation of imperfect boundary probability maps (BPM) in connectomics. Our algorithm can be a considered as an extension of spectral clustering. Instead of clustering the diffusion maps with traditional clustering algorithms, we learn the manifold and compute an estimate of the minimum normalized cut. We proceed by divide and conquer. We also introduce a novel criterion for determining if further splits are necessary in a component based on it's topological properties. Our algorithm complements the currently popular agglomeration approaches in connectomics, which overlook the geometrical aspects of this segmentation problem. 
\end{abstract}

\section{Introduction}

The connectome of an organism is the 3D connectivity graph of neurons and synapses that captures the structure of its nervous system \cite{sporns_human_2005}. As of today, there is only one animal, C. elegans to have its full connectome mapped \cite{brenner_genetics_1974}. The difficulty of the problem lies in the numbers. The manual segmentation of 300 neurons and 7000 synapses of C. elegans took about 12 years of part-time work by one scientist. To map the full human connectome, we would need to build a network of hundred billion neurons and about a thousand times as many synapses \cite{helmstaedter_2013}. 

Recently, there has been a great interest in developing a fully automated algorithms to reconstruct the connectome from high resolution 3D EM image stacks of the brain \cite{saturated_2015,automated_2015,lichtman_big_2014}. As described in Figure 8 of \cite{automated_2015}, the images-to-graph pipeline consists of several units. In this paper, our focus will be to improve the neuron segmentation unit. The task of neuron segmentation requires us to trace, or in other words color, the dendrites and axons of each neuron across the dense 3D volume.

Right now, GALA, the most successful algorithm for neuron segmentation combines three different machine learning algorithms: deep neural networks to perform boundary detection, watershed algorithm to generate an oversegmentation and random forests for agglomeration \cite{iglesias_machine_2013}. An example run of the algorithm is illustrated by transitions (a), (c) and (d) in Figure 1. In this paper, we attempt to improve this algorithm using Spectral Clustering. Our intuition tells us that while the BPM is imperfect, it only contains a few errors and there should be a way to take advantage of the geometry of the image to a greater extent. We will see later, that if we just threshold the BPM and find the connected components in this binary image, we already get a good number of the objects colored correctly.

The paper is organized as follows: in section 2, we introduce spectral clustering and the issue of this approach, in section 3 we outline the Hierarchical Manifold Spectral Clustering (HMSC) algorithm to remedy these issues, in section 4 we discuss our results and possible future directions.

\begin{figure}[h]
  \label{Flowchart}
  \includegraphics[width=4.8in]{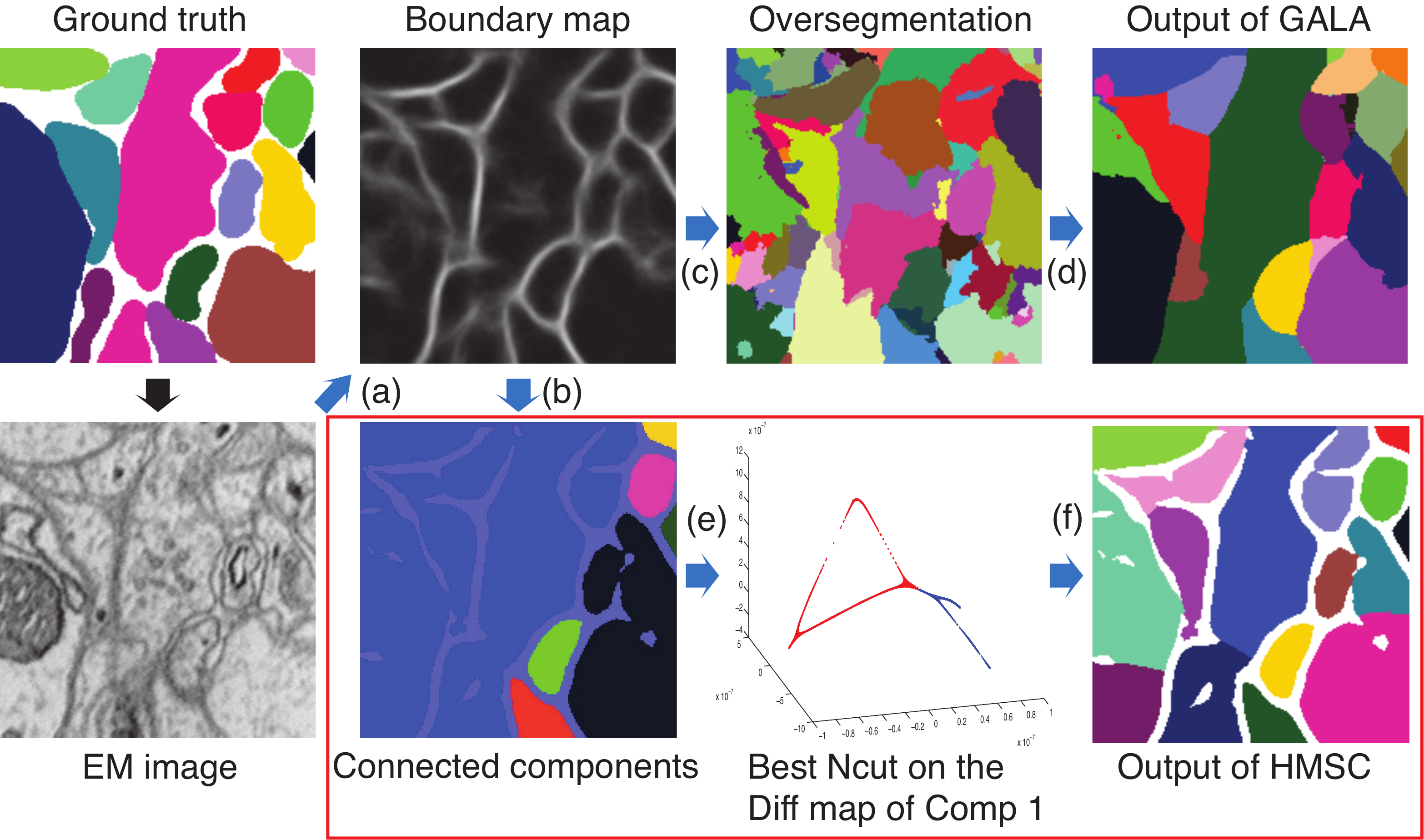}
  \caption
   {The different steps of neuron segmentation in GALA and in the newly developed HMSC algorithm (in red box). Each blue arrow indicates a set of algorithms. Arrow (a) indicates deep neural networks. (b) indicates finding the connected components. (c) indicates the watershed algorithm. (d) indicates the agglomeration algorithm in GALA. (e) indicates finding the minimum normalized cut in the diffusion map of one component. This can be thought of one divide step in the divide and conquer implemented in the HMSC algorithm. (f) indicates running divide and conquer until all components are determined to be correctly segmented by our stopping criterion. For comparison, we note that GALA ran on a $1024\times 1024\times 100$ dataset whereas HMSC ran on the $200\times 200$ Boundary map (see in figure), and we are showing the corresponding $200 \times 200$ image segments from the output of the two algorithms.}
\end{figure}

\section{Spectral Clustering}

\subsection{The algorithm from the textbook}

Spectral clustering is a popular approach for clustering a graph $G=(V,E)$ into $k$ components. Instead of clustering on the graph, the algorithm first embeds $G$ into $\mathbb{R}^{k-1}$ using a dimension reduction technique called diffusion maps, and then it proceeds by k-means clustering in $\mathbb{R}^{k-1}$. Intuitively, computing the diffusion map is equivalent to performing a random walk (or diffusion) from every point of the graph, and thereafter mapping the points into $\mathbb{R}^{d}$ based on the closeness of the probability distribution of their position after $t$ steps. This way the clusters in $G$ move closer together and become separated from other clusters.

More formally, let $A$ be the adjacency matrix, $D$ be the diagonal degree matrix, and $L=I-D^{-\frac12}AD^{-\frac12}$ be the symmetric normalized Laplacian matrix of $G$. Furthermore, let the eigenvectors and the corresponding eigenvalues of $L$ be denoted by $v_i$ and $\lambda_i$, and let $\varphi_i=D^{-\frac12}v_i$. Then, as defined in \cite{afonso_sc}, the diffusion map ${\phi}_t: V\rightarrow\mathbb{R}^{d}$ for node $V_i$ is 

\begin{equation}
\phi_t(V_i)=\left[
\begin{array}{c}
\lambda_2^k\varphi_2(i)\\
\lambda_3^k\varphi_3(i)\\
\vdots\\
\lambda_{d+1}^k\varphi_{d+1}(i)
\end{array}
\right] \ .
\end{equation}  
The indexing goes from $2$ to $d+1$ because the first eigenvector $\varphi_1$ of $L$ is always the all ones vector and does not contain any information. Note that we need to choose parameter $t$. According to our experiments, increasing $t$ didn't have significant impact on our mapping (data not shown), and for the rest of this paper we chose $t=1$.

As stated above, once the diffusion maps are calculated for $d=k-1$, spectral clustering proceeds with k-means clustering. We will not go into details about k-means here, for further information please refer to \cite{lloyd_least_2006} or any introductory machine learning textbook. We do note however, that choosing the parameter $k$ may not be obvious, and as we will see later the algorithm is not robust for changes in $k$.

\subsection{Applying spectral clustering to connectomics}

To apply spectral clustering to connectomics, we first need to preprocess the data; we have to extract a graph out of our input image. We define the nodes of our graph to be pixels with BPM value under a certain threshold. The threshold we manually optimized and was set to be 60 (pixels values were in range $[0,255]$). Note that as opposed to GALA, this way we have uncolored boundary pixels in our output coloring. To define edges, we connected each non-boundary pixel to all other non-boundary pixels inside their 8-connected neighborhood. Segmentation in connectomics is always in 3D, but for simplicity and speed we only considered 2D inputs in this paper. Nevertheless, we note that our algorithm simply generalizes to 3D by using 26 instead of 8-connectedness. Depending on how accurate the BPM was, the preprocessing might produce a disconnected graph. We cluster each connected component separately with a different $k$.

The algorithm was implemented in MATLAB with the built-in k-means cluttering algorithm. For seeding, we chose to use a preliminary clustering phase on random 10\% subsample of $G$. To test the performance of spectral clustering, we first created a test case of three clusters with five errors in the BPM. Figure 2 shows the input image, the diffusion map and the coloring produced by the algorithm with parameter $k=3$. In this case the coloring was perfect, but we report that due to the random seeding of k-means, in certain cases we did get an imperfect coloring (data now shown). After this encouraging first experiment, we investigate the effect of an imperfect $k$. In particular, we are most interested in what happens if we overshoot $k$ (so that we can perform agglomeration later if necessary). The results of spectral clustering on the same input image with parameter $k=4$ are shown on Figure 3. Surprisingly, the output coloring contains disconnected clusters. The understand the reason behind this phenomenon, remember that for $k=4$ the diffusion map has dimension three, so Figure 3 (b) is only a projection of the new data points. Image (c) in Figure 3 shows the points 3D from a different angle, which sheds light onto why output images like Figure 3 (d) are possible.

\begin{figure}[h]
  \label{Test3_good}
  \includegraphics[width=4.8in]{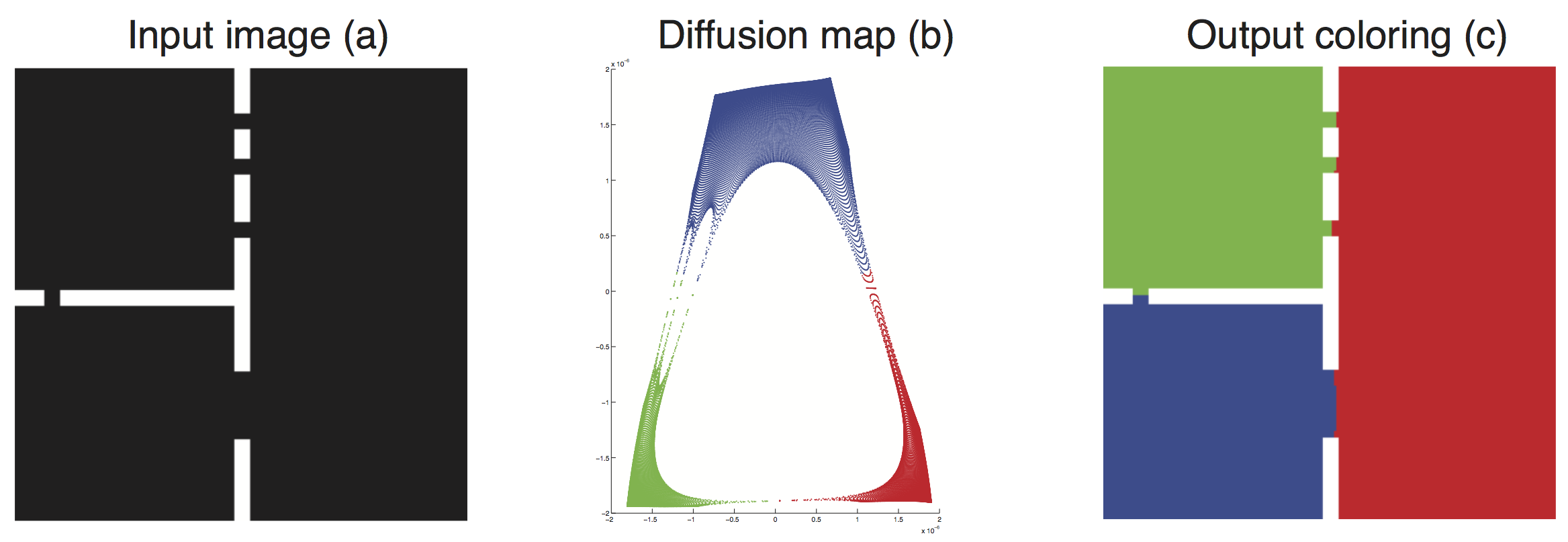}
  \caption
   {Spectral clustering on test input with $k=3$. Color coding in (b) and (c) are identical.}
\end{figure}

\begin{figure}[h]
  \label{Test3_bad}
  \includegraphics[width=4.8in]{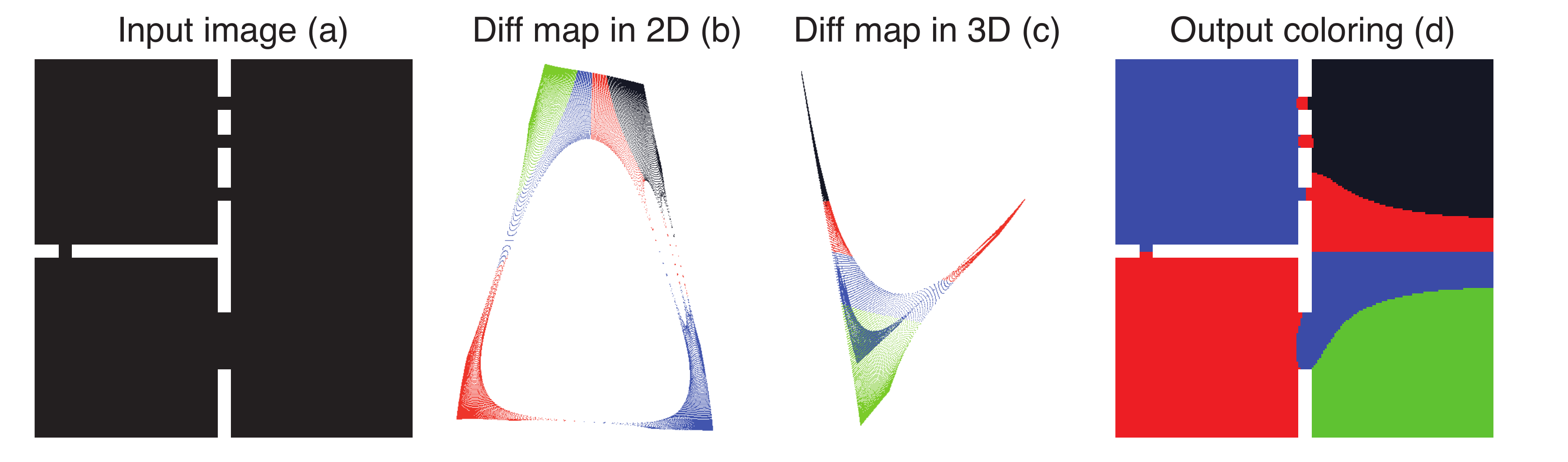}
  \caption
   {Spectral clustering on test input with $k=4$. Color coding in (b), (c) and (d) are identical.}
\end{figure}

This issue of outputting disconnected segments is very alarming especially in the connectomics setting, where learning the connectivity is the number one objective. One resolution of this issue would be split the segments even further based on their connectivity in the original graph G. This is acceptable, since our goal is to produce and oversegmentation. In this paper we follow an alternative approach.

We build intuition about the problem by running the algorithm above on a real biological image. Images (b) and (c) in Figure 4 show the input BPM and the connected components in our graph. Notice that some of the simple objects are segmented perfectly by this preprocessing step. To determine $k$ for each cluster we used a simple heuristic; we set $k_i=\lfloor\frac{\sqrt{n_i}}{2}\rfloor+1$, were $n_i$ is the number of nodes in component $i$. Similarly to the test dataset, the issue of disconnected clusters can be observed in our output in Figure 4 (d). However, the most important observation in Figure 4 comes from the diffusion maps (e) and (f). The two pairs of images show the diffusion map of a perfectly and an imperfectly segmented object of similar sizes. The topology of the two objects look very different. Loosely speaking, imperfectly segmented objects have diffusion maps with one dimensional internal structure, while the diffusion map of a perfectly segmented object tends to be two dimensional. We observed this trend in all components of this input image.

\begin{figure}[h]
  \label{Sp_real}
  \includegraphics[width=4.8in]{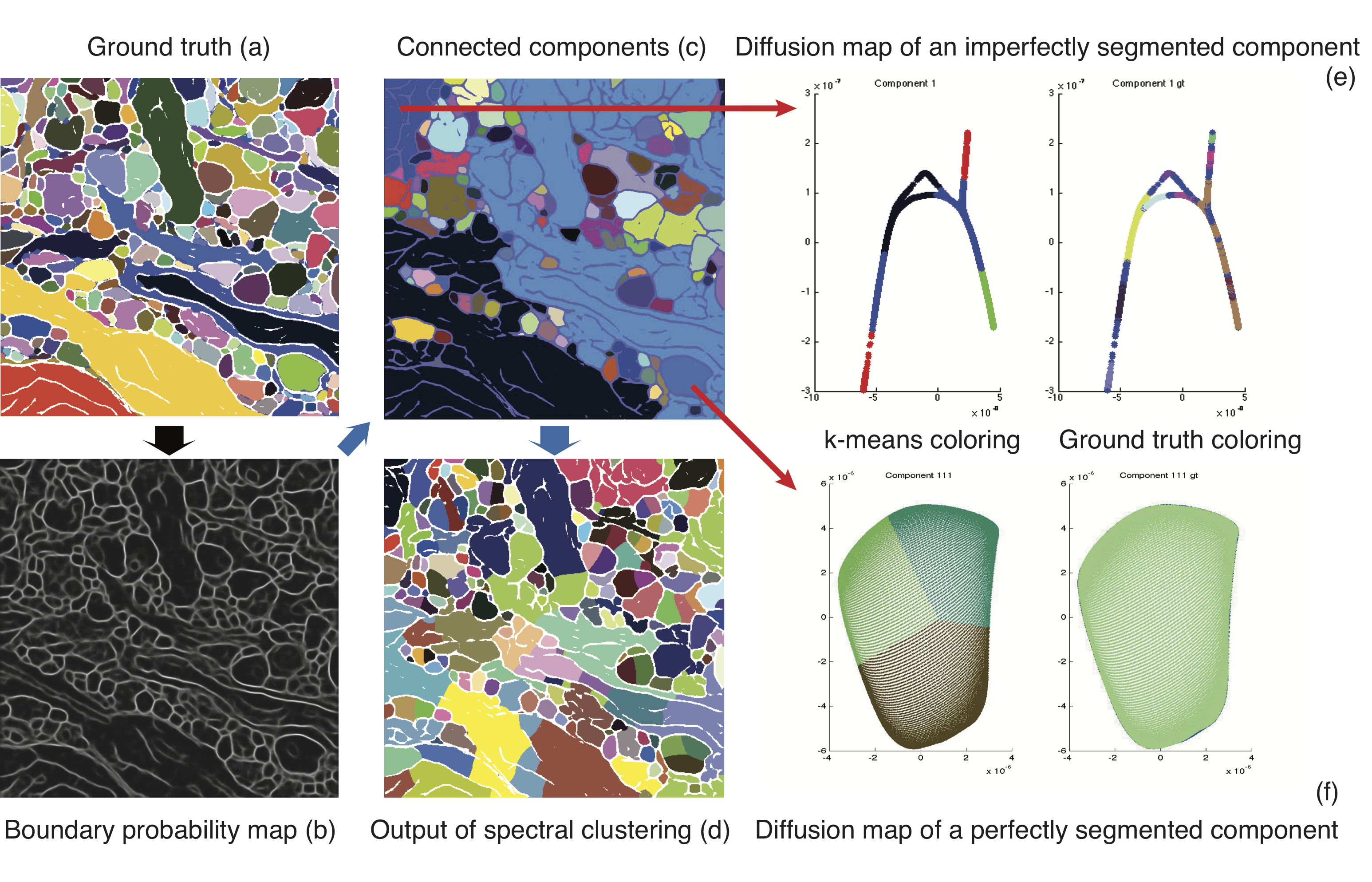}
  \caption
   {Spectral clustering on real biological data. Images (e) and (f) each show a pair of 2D diffusion maps. The red arrows indicate which connected components are shown. In both (e) and (f) the image on the right shows the coloring determined by k-means and the image on the left shows the true coloring. We used the same color coding in (d) and the left images of (e) and (f), and similarly in (a) and the right images of (e) and (f).  }
\end{figure}

\section{The HMSC algorithm}

Using the intuition from the previous section, we designed an algorithm which decides if an component is perfectly segmented or not, and if not it splits it into two components until necessary. The decision is made using the 3D diffusion map of the component. We chose 3D because that is the highest dimension we can easily visualize, but our algorithm generalizes to any number of dimensions above one. For splitting the component into two, we approximate its diffusion map with a truly one dimensional tree object, and compute the minimum normalized cut (see equation (2)) by trying all possible cuts. Here we take advantage of the 1D inner structure of the diffusion map, the tree approximation greatly reduces the number of possible cuts to check. The intuition behind this approach comes from Algorithm 3.2 in \cite{afonso_sc}. For better performance we compute a coarsened tree object with only about 200 nodes. In the rest of this section we discuss how to build this coarse tree approximation and how we compute the normalized cuts in the original image efficiently.

\tikzstyle{decision} = [diamond, draw, fill=blue!20, 
    text width=4.5em, text badly centered, node distance=3cm, inner sep=0pt]
\tikzstyle{block} = [rectangle, draw, fill=blue!20, 
    text width=5em, text centered, rounded corners, minimum height=4em]
\tikzstyle{block2} = [rectangle, draw, fill=green!20, 
    text width=5em, text centered, rounded corners, minimum height=4em]
\tikzstyle{line} = [draw, -latex']
\tikzstyle{cloud} = [draw, ellipse,fill=red!20, node distance=3cm,
    minimum height=2em]
 
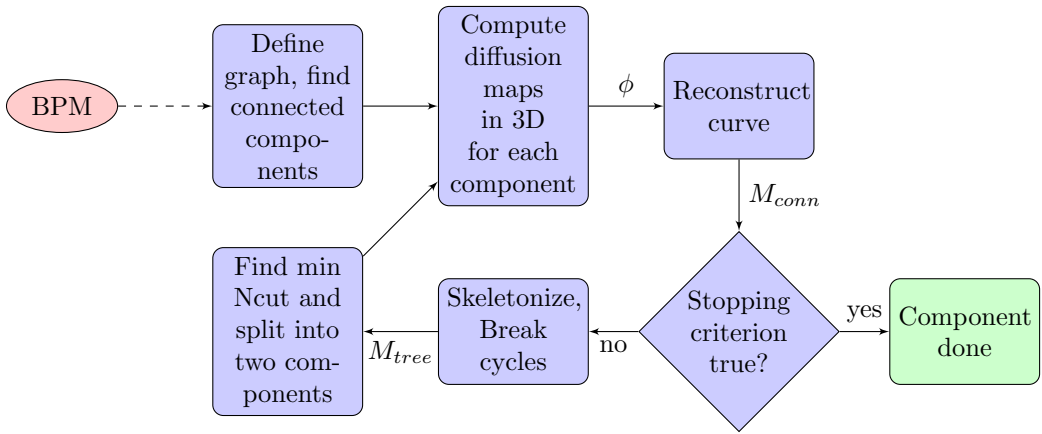
\begin{figure} [h]
\begin{tikzpicture}[node distance = 3cm, auto]
    \node [block] (conn_comp) {Define graph, find connected components};
    \node [cloud, left of=conn_comp] (BPM) {BPM};
    \node [block, right of=conn_comp] (diff_map) {Compute diffusion maps in 3D for each component};
    \node [block, right of=diff_map] (rec) {Reconstruct curve};
    \node [decision, below of=rec] (stop) {Stopping criterion true?};
    \node [block, left of=stop] (skel) {Skeletonize, Break cycles};
    \node [block, left of=skel] (split) {Find min Ncut and split into two components};
    \node [block2, right of=stop] (done) {Component done};

    \path [line]  (conn_comp) -- (diff_map);
    \path [line]  (diff_map) -- node {$\phi$} (rec);
    \path [line] (rec) -- node {$M_{conn}$} (stop);
    \path [line] (stop) -- node {yes} (done);
    \path [line] (skel) -- node {$M_{tree}$} (split);
    \path [line] (stop) -- node {no}(skel);
    \path [line] (split) -- (diff_map);
    \path [line,dashed] (BPM) -- (conn_comp);
\end{tikzpicture}
      \caption  {The flowchart of the HMSC algorithm. The algorithm is finished when all components are done.}
\end{figure}

\subsection{Coarsening and curve reconstruction}

We coarsen the diffusion map by embedding it into a $25 \times 25 \times 25$ image $M$. To keep track of which points were mapped to which cell in $M$ we define the list of sets $S_{x,y,z}=\{ \text{nodes in $V$ that were mapped to cell $(x,y,z)$ in $M$} \}$. We now face the challenge that $M$ might not be connected, especially if the nodes in the diffusion map are very well separated. The problem of connecting the points of a disconnected curve is referred to curve reconstruction in the literature. There are many algorithms available for this problem, for example \cite{dey_simple_1999}, but we have an extra advantage that we know which nodes were connected in the original graph $G=(V,E)$. Once we find an edge $e_{i,j} \in E$ that connects two disjoint segments in $M$, we compute the discretized version of the line $[\phi(V_i),\phi(V_j)]$ and add these points to $M'$. The most popular algorithm for computing such a discretized line is called Bresenham's line algorithm (see Figure 6). We used a MATLAB implementation of this algorithm developed by \cite{bresenham3d}. This algorithm produces a connected volume $M_{conn}$. 

\begin{figure}[h]
  \label{bresenham}
  \includegraphics[width=4.8in]{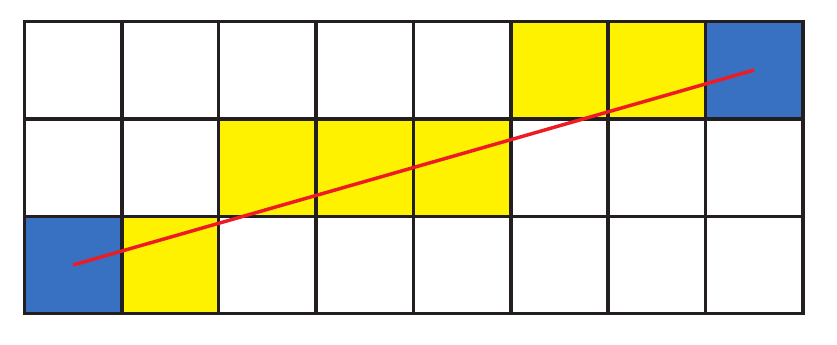}
  \caption
   {Illustration of Bresenham's line algorithm. The two blue squares are the input and the discretized line of yellow squares are the output of the algorithm. }
\end{figure}

\subsection{Stopping criterion}

At this point we use our intuition from the end of Section 2, to determine if the component should be split any further. To capture the dimensionality difference between perfect and imperfect segmentations, we tried setting the criterion to be a thresholding rule based on the average degree in the component. While this approach did work to some extent, we noticed that we achieve better results by considering the density of the coarsened image. We define the density $\delta_{x,y,z}=| S_{x,y,z} |$ for a cell to be the number of graph nodes mapped to that cell. Figure 7 (b) shows the density plot of the original volume $M$.
We found that the standard deviation of densities in imperfect segmentations are much higher than in perfect segmentations. For our criterion we used a threshold of 10 to determine if a component should be split any further.

\subsection{Skeletonization}

Skeletonization algorithms are designed to compute the midline of an object. We applied the most simple type of skeletionzation algorithm called a thinning algorithm to $M_{conn}$, to get a truly one dimensional object. Intuitively, the algorithm finds the boundary pixels of the 3D volume and deletes them as long as they don't  change the local topology of the object. This erosion process continues until the boundary shrinks into only a line, $M_{skel}$. Figure 7 (c) shows the skeletonized version of diffusion map shown in (a).

In this project, we used the MATLAB implementation \cite{Skeleton3D} of the thinning algorithm described in \cite{kerschnitzki_architecture_2013}. To maintain our mapping $S$ defined above, we modified \cite{Skeleton3D} so that each time cell $(x,y,z)$ is deleted, we find the closest cell still in the object $(x',y',z')$ and we set $S_{x',y',z'}=S_{x',y',z'} \cup S_{x,y,z}$ and $S_{x,y,z}=\emptyset$.

\subsection{Cycle breaking}

While skeletonization produces a one dimensional object, we still need to break the cycles go get a tree approximation. To break the cycles, we designed a heuristic using the density of each cell in $M_{skel}$. We perform cycle breaking by finding the union of all cycles in the graph, performing a random walk using the density of each cell as out initial state, and deleting the cell with the lowest probability after ten steps. When deleting a cell, we again maintain our mapping $S$ as we did during skeletonization. We perform the steps above until we get a tree, $M_{tree}$. The random walk is necessary because due to the previous steps in the algorithm, it might be possible that there is a low density cell between two high density cells and the random walk evens out these artifacts. 

\subsection{Computing minimum normalized cut}

Let the tree graph induced by $M_{tree}$ be $G_t=(V_t,E_t)$. Once we have the coarsened tree approximation of the diffusion map, for each edge in $e \in E_t$ we compute the value normalized cut in $G$ induced by that cut $E_t \setminus \{e \}$. The normalized cut for $W\subset V$, $W\ne \emptyset$ as defined in \cite{afonso_sc} is   

\begin{equation}
Ncut(W) = \frac{Cut(W)}{Vol(W)}+\frac{Cut(W^{C})}{Vol(W^{C})}\ ,
\end{equation} 
where if $\mathcal{X}_G$ is the indicator function of edges in $G$,
\begin{equation}
Cut(W) = \sum_{v_i\in W, v_j\in W^{C}} \mathcal{X}_G(v_i,v_j)\ .
\end{equation} 
Since our graph is undirected, in our case $Cut(W^{C})=Cut(W)$ and $Vol(W)=|W|$.

To speed up our algorithm, before computing the normalized cuts, we build an extended adjacency graph of $G_t$; let $G_{EAG}=(V_{EAG}, E_{EAG})$, where $V_{EAG}= V_t$ and for $v_i, v_j \in V_{EAG}$, let $e_{v_1,v_2}=\sum_{i,j} \mathcal{X}_G(S_{v_1}(i),S_{v_2}(j))$, where $S$ is still our inverse mapping from cells to graph nodes.

If we cut edge $e_t\in E_t$, the set $V_t$ falls into two components $C_{e_t}$ and $C_{e_t}^{C}$. Then the value of our cut will be

\begin{equation}
Cut(C_{e_t})= \sum_{v_i\in C_{e_{t}}, v_j\in C_{e_{t}}^{C}} e_{v_1,v_2}\ .
\end{equation}
The power of the Extended Adjacency Graph, is that this way we only need to sum over the edges of the $G_{EAG}$ instead of the edges of $G$, and $|E_{EAG}| << |E|$. The volume $Vol(C_{e_t})$ in our case just equals to the sum of the densities $\sum_{v_i\in C_{e_{t}}}\delta_{v_i}$.

We note however, that if we look at the inverse image of $C_{e_t} \subset V_t$ in the original map $G$, we do not need to get a connected component due to the coarsening. Fortunately, since the tree approximation describes the diffusion map very accurately, the disconnected part can only be minor. In fact we can join the small disconnected part to the other side of the cut and the value of the normalized cut will change only by little. 

Once $Ncut(C_{e_t})$ is calculated for all $e_t \in E_t$, we split the current component along the cut corresponding to the minimum normalized cut. The algorithm proceeds recursively until all components satisfy the stopping criterion.

\begin{figure}[h]
  \label{Fig6}
  \includegraphics[width=4.8in]{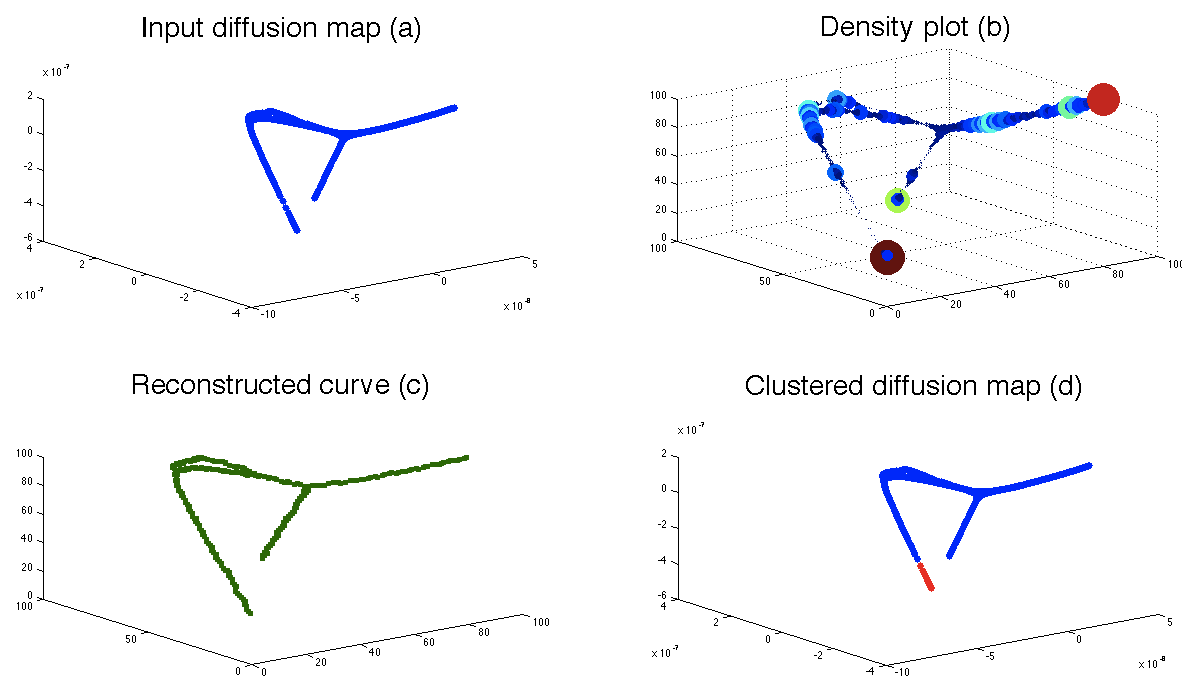}
  \caption
   {The progression of the HMSC algorithm on one particular diffusion map. Image (a) shows the 3D diffusion map of the component. Image (b) shows the density plot of the coarsened image $M$. Larger spheres with warmer color imply higher density. Image (c) shows diffusion map (a) after coarsening, curve reconstruction and skeletonization. Image (d) shows the diffusion map (a) with the coloring along the minimum normalized cut.}
\end{figure}

\section{Discussion}
We presented an alternative approach for neuron segmentation with imperfect BPMs. Unfortunately, our implementation is still rather slow (the example shown in the red box of Figure 1 took about two minutes to run) and we were unable to compare our result with GALA's result using any of the popular segmentation metrics in connectomics.

\subsection{Runtime analysis}
In the divide step our algorithm is linear in the number of pixels $n$. If we ensure that we only consider cuts with $min(Vol(W), Vol(W^C))>\frac{n}{10}$, then our algorithm will have the recurrence

\begin{equation}
T(n)< 2*T(9/10)+O(n)\ ,
\end{equation}
which implies a time complexity of $O(n\log n)$. The fact that our algorithm is still slow is probably due to implementation inefficiencies.
\subsection{Further research directions} 

The algorithm we presented in this paper is quite complex and introduces several new ideas. It is possible that a subset of these ideas will help improve current segmentation techniques in connectomics. Since the algorithm is currently too slow, it might be interesting to use some heuristic shortcuts that only take advantage of some of the new techniques introduced above. 

For example, we could further separate the disconnected clusters returned by k-means as described in Section 2. Alternatively, we could define a normalized cut that only depends on the density of $M_{tree}$, which would save us from computing the extended adjacency graph (one of the more expensive steps). 

Our most striking discovery was observing the topological differences of perfectly and imperfectly segmented components, however, the stopping criterion that we implemented only considers the density of the diffusion map. Further research on more advanced stopping criteria is necessary.

This paper mainly focuses on a new algorithm and its application, however, we encountered theoretical questions along the way. We identify a possible theoretical research direction of investigating that under what circumstances is the topological dimension of a diffusion map small.

\section{Acknowledgement}

This paper was the final project for the class 18.S096: Topics in Mathematics of Data Science in Fall 2015 at MIT. The author is grateful for Professor Afonso Bandeira for his supervision on this project. The author would also like to thank members of the MIT Computational Connectomics Group for making their nonpublic boundary prediction maps available for this class project.

\bibliographystyle{plain}
\bibliography{final}

\end{document}